\documentclass[runningheads]{llncs}

\usepackage{eccv}

\usepackage{eccvabbrv}
\usepackage[table,xcdraw]{xcolor}
\usepackage{graphicx}
\usepackage{booktabs}
\usepackage{multirow}
\usepackage{placeins}
\usepackage{pdflscape}   % 如果要整页横向
\usepackage{afterpage}   % 如果要强制下一页开始
\usepackage{placeins}
\usepackage{marvosym}

\usepackage[accsupp]{axessibility}  % Improves PDF readability for those with disabilities.

\usepackage[pagebackref,breaklinks,colorlinks,citecolor=eccvblue]{hyperref}

\usepackage{orcidlink}

\usepackage{graphicx}
\usepackage{subcaption}

\usepackage[utf8]{inputenc}
\begin{document}

% ---------------------------------------------------------------
% TODO: Replace with your title
\title{Attention-DP3: Spatially Object-aware 3D Diffusion Policy via Geometry-aligned Attentional Conditioning}

% TODO: If the paper title is too long for the running head, you can set an abbreviated title here.
\titlerunning{Attention-DP3}

\author{Changbo Yan$^{*}$, Zhongbo Zhang$^{*}$, Zaibin Zhang$^{*}$, Lijun Wang, Yifan Wang\textsuperscript{\Letter}, \\  Huchuan Lu}

\authorrunning{C.~Yan et al.}

\institute{ Dalian University of Technology,
\\
\texttt{wyfan@dlut.edu.cn}
\\
$^*$ Equal contribution
\qquad
\textsuperscript{\Letter} Corresponding author
% $^\dagger$ Project leader
}

% % TODO: Replace with your author list (keep anonymized for review if needed).
% \author{Anonymous ECCV Submission}

% % TODO: Abbreviated list of authors for the running head.
% \authorrunning{Anonymous}

% % TODO: Replace with your institution list for camera-ready.
% \institute{}

\maketitle

\begin{abstract}
3D point-cloud observations are inherently ambiguous in complex, cluttered manipulation scenes, where target objects may be partially occluded or tightly intermingled with visually similar distractors. As a result, standard 3D diffusion policies often struggle to localize and exploit task-relevant geometry as scene complexity grows.
We propose \textbf{Attention-DP3}, a spatially object-aware 3D diffusion policy that injects object-level geometric cues via attention while keeping the DP3 diffusion backbone unchanged. Our pipeline performs open-vocabulary 2D segmentation on RGB images, then lifts predicted target masks into 3D using calibrated camera geometry to obtain object-centric geometric priors.
We incorporate these cues through Tri-field Attentional Conditioning, which constructs three complementary fields: (i) a targetness field to anchor the target object, (ii) an intra-target saliency field to emphasize task-relevant geometry within the target, and (iii) a backgroundness field to suppress distractors and clutter.
Experiments on Adroit, DexArt, MetaWorld, and the real-world SO101 platform show consistent improvements over DP3, achieving state-of-the-art performance across benchmarks. Notably, as distractor objects increase, DP3 drops sharply, whereas Attention-DP3 remains stable and outperforms DP3 by up to 31\% under heavy clutter. The code is publicly available at \url{https://github.com/zhangzhongbo2213/Attention-DP3}.


\end{abstract}

% We introduce \textbf{Attention-DP3}, a semantically grounded 3D diffusion policy to mitigate perceptual ambiguity in cluttered manipulation.
% Our method leverages open-vocabulary 2D grounding and segmentation to obtain target masks from RGB images, and \emph{explicitly lifts} them into the 3D point cloud via calibrated camera geometry.
% We construct a deterministic \emph{Tri-Perspective Attentional Conditioning} signal that provides (i) target anchoring, (ii) structural saliency shaping, and (iii) distractor decoupling, and train only a lightweight attention encoder together with a standard DP3-style diffusion policy.
% Across Adroit, DexArt, and MetaWorld, Attention-DP3 consistently outperforms the DP3 baseline and shows strong robustness under unseen visual clutter.
%   \keywords{Diffusion Policy \and 3D manipulation \and semantic attention}
% \end{abstract}

% Main paper body. Supplementary material is compiled separately from sup.tex.
\section{Introduction}
\label{sec:intro}

\begin{figure}[t]
    \centering
    \includegraphics[width=0.8\columnwidth]{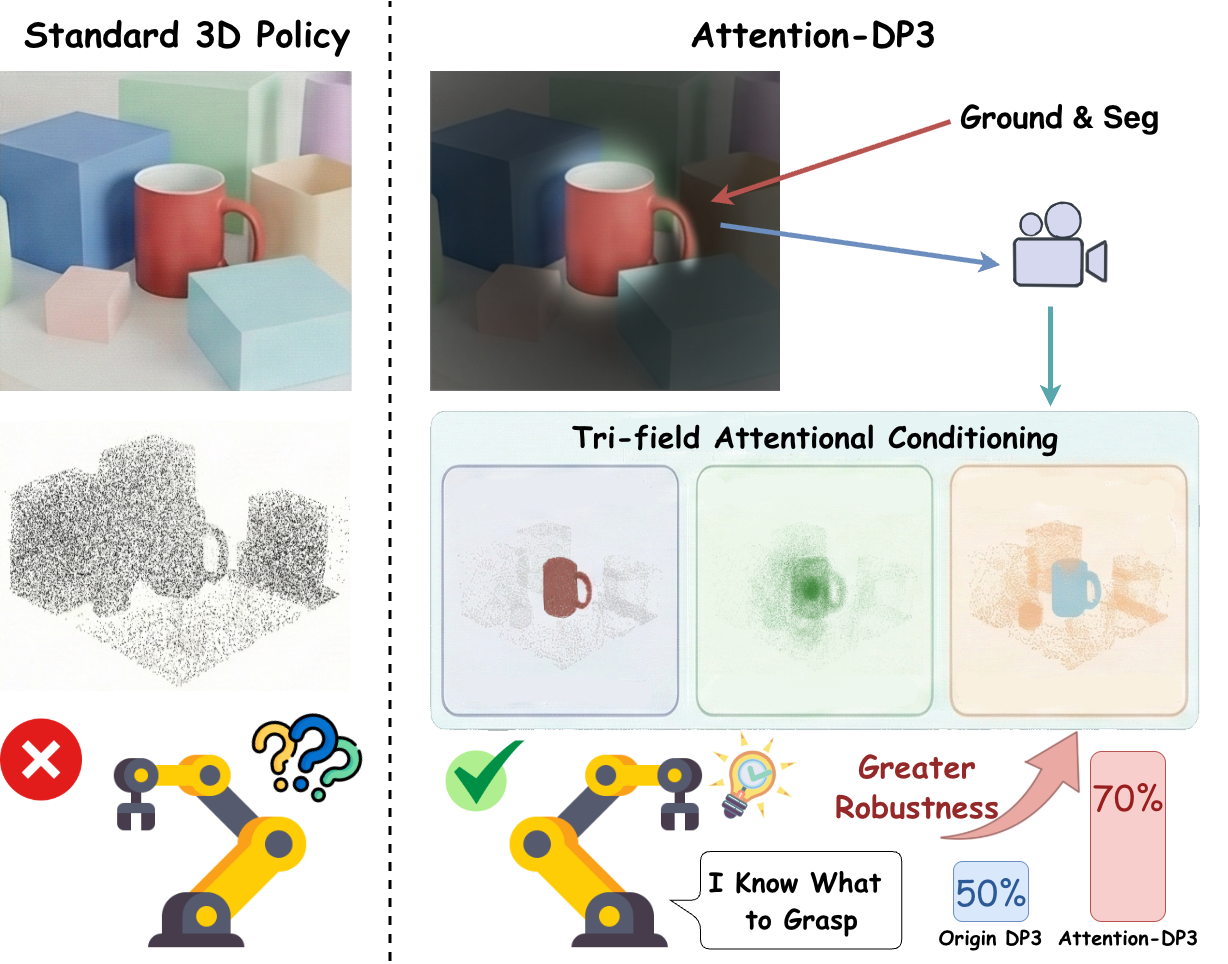}
    \caption{\textbf{Standard 3D policies vs. Attention-DP3.}
    \textbf{(Left)} In cluttered scenes, sparse point clouds can be perceptually ambiguous under occlusion and distractors, causing unreliable target association.
    \textbf{(Right)} Attention-DP3 lifts object masks from RGB into 3D and uses geometry-aligned, object-aware attentional conditioning to guide action generation, helping the policy focus on what to grasp and where it is in 3D.}
    \label{fig:teaser}
\end{figure}

Recent progress in imitation learning (IL)~\cite{zare2024survey,chi2025diffusion,black2024pi_0,black2025pi_,kim2024openvla,brohan2022rt,zitkovich2023rt} has enabled robot policies to solve complex manipulation tasks with improved stability and generalization.
In particular, diffusion-based policies~\cite{chi2025diffusion,team2024octo,park2024diar,liang2024skilldiffuser} generate actions via iterative denoising, providing an expressive and robust framework for multimodal behavior synthesis.
Recent 3D diffusion variants~\cite{ze20243d,ke20243d,li2025language,lu2024manicm,ke2024bi3d} further improve generalization by conditioning on point clouds, leveraging explicit geometry to better handle viewpoint changes and out-of-distribution object configurations.

Despite these advances, manipulation in unstructured environments remains bottlenecked by perception~\cite{ze20243d}.
While point clouds provide metric spatial constraints, they are sparse and highly ambiguous under clutter: when targets are partially occluded or tightly intermingled with distractors, the observed 3D points no longer cleanly separate the object of interest from background.
This ambiguity induces a recurring failure mode for 3D diffusion policies, where the policy cannot reliably bind the language-specified target to its 3D extent, leading to unstable grasps or actions drifting toward distractors as clutter increases.

A natural direction is to exploit object-level cues from RGB observations to disambiguate the 3D scene~\cite{shen2023distilled,ze2023gnfactor,gu2024conceptgraphs,peng2023openscene,huang2023voxposer}.
However, directly fusing dense RGB features with sparse point-cloud representations can be brittle in clutter and may compromise geometric precision.
Instead, we treat object-level cues as lightweight attentional prompts that modulate point-cloud-conditioned policy learning while preserving the underlying 3D diffusion backbone.

We propose \textbf{Attention-DP3}, a new 3D diffusion policy model that resolves clutter-induced perceptual ambiguity via object-aware 3D attention, while keeping the DP3 diffusion formulation unchanged.
Given a language goal, our pipeline first performs open-vocabulary 2D segmentation on RGB images, then lifts the predicted target masks into 3D using calibrated camera geometry to obtain geometry-aligned object priors.
We introduce Tri-Field Attentional Conditioning, which decomposes object-aware guidance into three complementary fields over 3D points: (i) a targetness field that anchors target points, (ii) an intra-target saliency field that emphasizes task-relevant geometry within the target, and (iii) a backgroundness field that suppresses distractors while retaining sufficient context for occlusion reasoning.
These fields act as soft attention cues, enabling object-aware disambiguation without altering the diffusion backbone.

We evaluate Attention-DP3 on Adroit~\cite{rajeswaran2017learning}, DexArt~\cite{bao2023dexart}, MetaWorld~\cite{yu2020meta}, and the real-world SO101 platform, where it consistently improves over DP3.
Beyond average gains, we conduct stress tests that systematically scale visual clutter by increasing distractor objects at test time.
Notably, DP3 exhibits a sharp performance drop as clutter intensifies, whereas Attention-DP3 remains stable and achieves gains of up to +31\% under extreme distraction, highlighting strong zero-shot robustness to unseen clutter.

Our contributions are threefold.
\textbf{(1) Spatially object-aware prompting for 3D diffusion policies:} we lift open-vocabulary 2D object masks into 3D and use them as geometry-aligned attentional prompts for point-cloud-conditioned policy learning, without modifying the diffusion backbone.
\textbf{(2) Tri-Field Attentional Conditioning:} we propose a lightweight multi-field conditioning design decomposing object-aware guidance into target anchoring, intra-target structure emphasis, and distractor suppression for robust perception under clutter.
\textbf{(3) Consistent gains and strong clutter robustness:} we demonstrate improvements across multiple benchmarks and substantial zero-shot robustness under systematically increased unseen clutter, including extreme distraction scenarios.
The code is publicly available at \url{https://github.com/zhangzhongbo2213/Attention-DP3}.

\section{Related Work}
\label{sec:related_work}

\subsection{3D Representations for Robotic Manipulation}
Point cloud learning has become a standard approach for 3D robotic perception. Early works such as PointNet~\cite{qi2017pointnet} and PointNet++~\cite{qi2017pointnet++} processed raw point clouds directly with permutation-invariant architectures. Subsequent methods incorporate local geometry with convolution-style operators (e.g., KPConv~\cite{thomas2019kpconv}) or attention mechanisms (e.g., Point Transformer~\cite{zhao2021point}), improving feature expressiveness for 3D understanding. In manipulation settings, point-cloud encoders frequently serve as lightweight perception backbones within policies~\cite{ze20243d,qin2023dexpoint,shridhar2023perceiver}. Compared to 2D-only representations, explicit 3D geometry reduces viewpoint sensitivity and provides accurate spatial constraints for contact-rich control~\cite{gervet2023act3d,goyal2023rvt,ze2023gnfactor}. However, pure geometry lacks semantic grounding, motivating its combination with task-relevant semantic cues.

\subsection{Diffusion Policies for Imitation Learning}
Diffusion policy models action generation as an iterative denoising process, providing a stable and expressive framework for learning multimodal behaviors from demonstrations. Recent works extend diffusion policies to 3D observations by conditioning on point clouds, typically encoding them into compact latent vectors to guide a conditional denoiser~\cite{ze20243d,ke20243d,li2025language,lu2024manicm,ke2024bi3d}. This family of approaches has shown strong generalization across manipulation tasks and has inspired further improvements in robustness and scaling, including equivariant architectures (e.g., EquiBot~\cite{yang2024equibot}) and large-scale pre-training (e.g., FP3~\cite{yang2025fp3}).
Our method is built on this line of 3D diffusion policies, but augments the conditioning with a semantic prior that is explicitly aligned to the observed geometry, rather than learned through joint feature fusion.

\subsection{Semantic-Guided 3D Control}
Recent works increasingly use 2D foundation models to provide strong semantic priors for robotic manipulation~\cite{liu2023grounding, ravi2024sam}. However, integrating these 2D signals into 3D control or attention mechanisms typically relies on learned cross-modal feature fusion~\cite{jia2024lift3d} or explicit mask-based cropping~\cite{huang2025roboground, zhu2023vima}. These approaches can be brittle under visual clutter and often compromise strict spatial precision required for delicate tasks. Unlike methods requiring complex learned alignments or heavy pre-processing, we bypass 2D-3D feature fusion. We instead apply explicit geometric lifting to construct a training-free, geometry-aligned semantic attention signal, providing robust and lightweight guidance for 3D policies.
\section{Method}
\label{sec:method}

\subsection{Overview}
\label{sec:overview}
\begin{figure*}[t]
    \centering
    \includegraphics[width=1.0\textwidth]{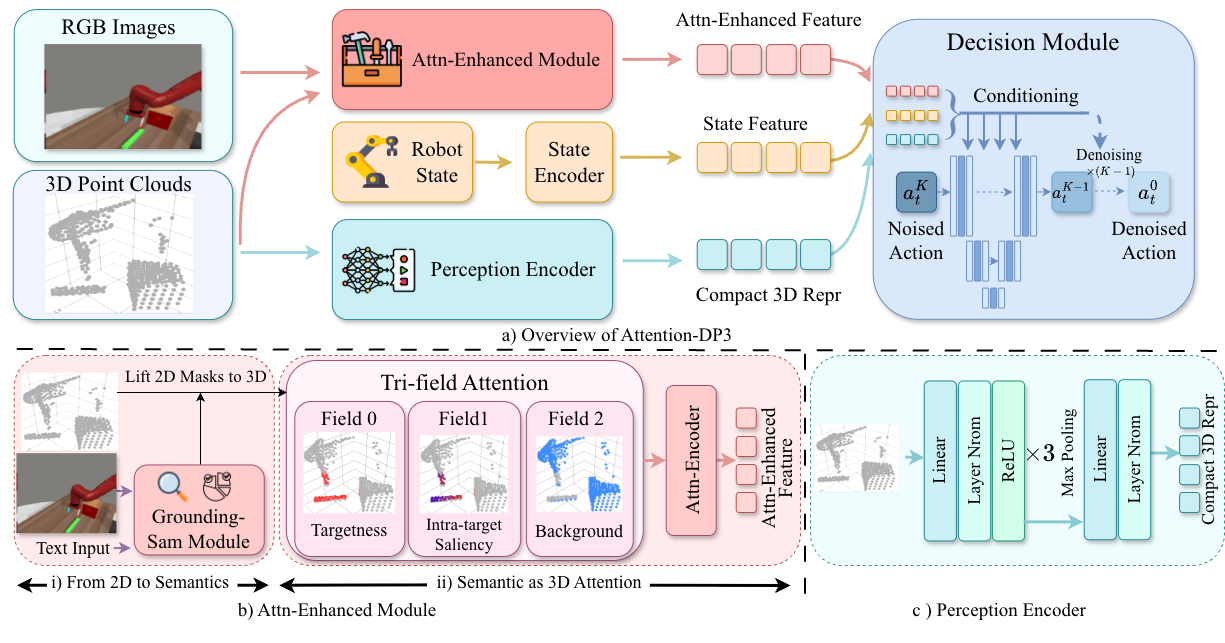}
\caption{\textbf{Attention-DP3 overview.}
\textbf{(a) Pipeline.} The policy conditions on synchronized RGB, 3D point clouds, and proprioception. A 3D Perception Module encodes global geometry, a State Encoder embeds proprioception, and an Attn-Enhanced Module maps open-vocabulary 2D cues to an attention feature. Their embeddings condition a diffusion Decision Module to denoise an action trajectory.
\textbf{(b) Attn-Enhanced Module.} A frozen grounding-and-segmentation pipeline predicts a text-conditioned 2D mask, which is lifted to three geometry-aligned fields: Targetness, Intra-target Saliency, and Backgroundness. These fields are individually encoded by Field Encoders and subsequently aggregated via an MLP.
\textbf{(c) Perception Encoder.} A lightweight point-cloud backbone encodes raw point clouds into a compact 3D representation.}
    \label{fig:overview}
\end{figure*}

As shown in Fig.~\ref{fig:overview}, Attention-DP3 injects language-conditioned object awareness into DP3 through a geometry-aligned tri-field attention.
At each timestep, the policy observes an RGB image $I$, a point cloud $P=\{p_i\}_{i=1}^{N}$, and proprioception $s$.
A 3D perception encoder extracts a compact geometry feature $z_{\mathrm{pc}}$, while a lightweight state encoder maps $s$ to $z_{\mathrm{state}}$.
In parallel, we design an Attn-Enhanced Module. A frozen grounding-and-segmentation pipeline (Grounding DINO + SAM2)~\cite{liu2023grounding,ravi2024sam} predicts a text-conditioned 2D mask on $I$, which we lift onto $P$ to obtain three geometry-aligned fields: Targetness, Intra-target Saliency, and Backgroundness.
These fields are individually encoded by field encoders and subsequently aggregated into an attention feature $z_{\mathrm{attn}}$ via an MLP.
The diffusion denoiser (conditional U-Net) is conditioned on
\begin{equation}
c = [z_{\mathrm{pc}},\, z_{\mathrm{state}},\, z_{\mathrm{attn}}],
\label{eq:cond_overview}
\end{equation}
and generates an $H$-step action trajectory via iterative denoising.
By grounding language cues in 3D fields instead of learning implicit RGB--3D alignment, the policy remains robust in cluttered scenes.
The tri-field design factorizes complementary roles: binding to the target (Targetness), emphasizing within-target structure (Intra-target Saliency), and suppressing distractors while retaining context (Backgroundness).

\subsection{Explicit 2D-to-3D Object-Cue Lifting}
\label{sec:semantic_lifting}
A central challenge in cluttered manipulation is that sparse point clouds alone often do not cleanly separate the object of interest from surrounding distractors, particularly under occlusion.
A common approach is to learn implicit RGB--3D alignment in a shared feature space and fuse representations across modalities.
However, this learned alignment can be brittle in heavy clutter and can blur geometry that is critical for precise control.
We adopt an explicit lifting strategy instead: a text-conditioned object mask is extracted in 2D and deterministically mapped onto the observed 3D points.
Given an RGB image $I\in\mathbb{R}^{H\times W\times 3}$ and a language query, a frozen segmentation model outputs a binary mask $M\in\{0,1\}^{H\times W}$ that marks pixels belonging to the queried object.
With calibrated camera intrinsics and extrinsics, each 3D point $p_i$ is projected onto the image plane:
\begin{equation}
(u_i, v_i) = \pi(p_i),
\label{eq:proj_rewrite}
\end{equation}
Here $\pi(\cdot)$ is the calibrated 3D-to-2D projection operator induced by the camera intrinsics and extrinsics: it maps a 3D point to its corresponding pixel coordinate under the pinhole camera model (with points behind the camera or outside the image treated as background).

For mask lookup, we discretize the continuous projection to the nearest pixel index.
Each point then receives a lifted object indicator through nearest-neighbor lookup:
\begin{equation}
m_i = M(u_i, v_i), \qquad m_i\in\{0,1\}.
\label{eq:lift_mask_rewrite}
\end{equation}

The resulting $\{m_i\}_{i=1}^{N}$ provides a geometry-aligned object cue over the point set: it is derived from RGB while remaining strictly consistent with the observed 3D geometry via calibrated projection.
This explicit lifting avoids learning 2D--3D correspondence in an embedding space and offers a robust bridge from RGB object cues to point-cloud policy conditioning.
We then refine the lifted cue into a tri-field attention representation to better handle clutter and occlusion.

\subsection{Geometry-aligned Lifted Tri-Field Attention}
\label{sec:attention_field}
From the lifted indicator $\{m_i\}$, we construct a deterministic Lifted Tri-Field Attention (LTFA) signal
$A\in\mathbb{R}^{3\times N}$.
LTFA appends three complementary per-point channels to the point cloud while leaving the 3D coordinates unchanged, thereby injecting object-aware attention without altering any geometric measurements.
For each point $p_i$, we define
$A_{\cdot,i}=[A_{\mathrm{T},i},A_{\mathrm{S},i},A_{\mathrm{B},i}]^\top$,
where each component is a scalar field designed to stabilize conditioning under clutter and occlusion.

\subsubsection{Field 1: Targetness.}
We define a targetness field that provides an explicit where-to-attend signal in 3D:
\begin{equation}
A_{\mathrm{T},i} = m_i.
\label{eq:ta_rewrite}
\end{equation}

\subsubsection{Field 2: Intra-target saliency.}
Binary masks can be noisy or fragmented and may miss thin parts under occlusion.
To provide a graded structural cue within the target region, we compute a normalized distance field
$D(u,v)\in[0,1]$ from the 2D mask geometry and lift it to points:
\begin{equation}
s_i = D(u_i,v_i), \qquad A_{\mathrm{S},i}= m_i \cdot s_i.
\label{eq:sss_rewrite}
\end{equation}
In practice, $D$ is obtained via the Euclidean distance transform inside the mask and normalized to $[0,1]$.
This field softly emphasizes geometrically central or contact-relevant subregions instead of uniformly weighting all target points.

\subsubsection{Field 3: Backgroundness.}
Rather than discarding non-target points, we encode complementary context via a backgroundness field:
\begin{equation}
A_{\mathrm{B},i} = 1 - m_i.
\label{eq:dd_rewrite}
\end{equation}
This encourages the downstream encoder to down-weight distractor evidence while preserving enough context for reasoning about occlusion and scene layout.

\subsubsection{Field encoder.}
While LTFA itself is deterministic and training-free, we learn a lightweight field encoder
$E_{\mathrm{field}}$ (point-wise MLP + symmetric pooling) to aggregate per-point fields into a compact vector:
\begin{equation}
z_{\mathrm{attn}} = E_{\mathrm{field}}(A).
\label{eq:zattn_rewrite}
\end{equation}
Treating $A$ as a three-channel per-point feature, we apply a shared MLP to each column $A_{\cdot,i}$ and pool across points, yielding a global attention feature. Grounding/segmentation models, mask lifting, and LTFA construction (Eqs.~\eqref{eq:proj_rewrite} to \eqref{eq:dd_rewrite}) remain frozen; only $E_{\mathrm{field}}$ and downstream DP3 components are learned.

\subsection{Attention-conditioned Diffusion Policy}
\label{sec:diffusion}
We follow the DP3 formulation and condition the denoiser on $c$ (Eq.~\eqref{eq:cond_overview}).
Let $\mathbf{x}_0\in\mathbb{R}^{H\times d_a}$ denote the clean action sequence (horizon $H$, action dimension $d_a$).
The forward diffusion process corrupts $\mathbf{x}_0$ with Gaussian noise, producing $\mathbf{x}_t$ at diffusion step $t\in\{1,\dots,T\}$:
\begin{equation}
q(\mathbf{x}_t \mid \mathbf{x}_0) =
\mathcal{N}\!\left(\mathbf{x}_t;\ \sqrt{\bar{\alpha}_t}\mathbf{x}_0,\ (1-\bar{\alpha}_t)\mathbf{I}\right),
\label{eq:forward_rewrite}
\end{equation}
where $\bar{\alpha}_t=\prod_{k=1}^{t}\alpha_k$ is determined by a fixed noise schedule $\{\alpha_t\}_{t=1}^{T}$.

The policy is parameterized by a conditional denoiser $\epsilon_\theta$.
Following DP3, we instantiate $\epsilon_\theta$ as a conditional U-Net over the temporally structured action sequence, where the diffusion step $t$ is embedded and injected into each residual block.
The global condition $c$ is incorporated via FiLM-style feature modulation in intermediate layers, enabling the denoiser to couple action generation with both geometry features and the proposed tri-field attention.
Formally, the denoiser predicts the injected noise:
\begin{equation}
\hat{\epsilon} = \epsilon_\theta(\mathbf{x}_t,\ t,\ c).
\label{eq:epspred_rewrite}
\end{equation}
We train with the standard noise-prediction objective:
\begin{equation}
\mathcal{L}(\theta) =
\mathbb{E}_{\mathbf{x}_0,t,\epsilon}\!\left[
\left\|\epsilon - \epsilon_\theta(\mathbf{x}_t,\ t,\ c)\right\|_2^2
\right],
\label{eq:loss_rewrite}
\end{equation}
where $\epsilon\sim\mathcal{N}(0,\mathbf{I})$ and $\mathbf{x}_t$ is sampled via Eq.~\eqref{eq:forward_rewrite}.
At inference, we start from $\mathbf{x}_T\sim\mathcal{N}(0,\mathbf{I})$ and iteratively apply $\epsilon_\theta$ from $t=T$ to $1$, yielding the final action trajectory $\mathbf{x}_0$.

\section{Experiments}
\label{sec:experiments}

\subsection{Experiment Setup}

\subsubsection{Benchmarks.}
Adroit~\cite{rajeswaran2017learning} contains highly challenging 24-DoF anthropomorphic hand manipulation tasks. DexArt~\cite{bao2023dexart} focuses on articulated-object manipulation, which requires precise spatial reasoning. MetaWorld~\cite{yu2020meta} provides diverse and structurally varied tabletop manipulation tasks, such as Pick-Place, Box-Close, and Shelf-Place. SO101 includes Place Cube, Push Cube, and Stack Cube. 

\subsubsection{Data Collection.}
We obtain expert demonstrations using a scripted policy for MetaWorld, and employ VRL3~\cite{wang2022vrl3} and PPO~\cite{schulman2017proximal} for the Adroit and DexArt environments, respectively. Our dataset includes 10 episodes per task for the Adroit and MetaWorld benchmarks, and 100 episodes for DexArt. For SO101, we collect 10 expert trajectories via LeRobot teleoperation using a global camera at 30 Hz; each SO101 arm has 6 degrees of freedom.

\subsubsection{Baselines.}
We evaluate Attention-DP3 against a wide spectrum of state-of-the-art visuomotor policies, which can be categorized into three groups: 
(i) Implicit Behavioral Cloning Baselines: including BCRNN~\cite{mandlekar2021matters} and Implicit Behavioral Cloning (IBC)~\cite{florence2022implicit}; 
(ii) Image-based Generative Policies: including the foundational Diffusion Policy (DP)~\cite{chi2025diffusion}, Consistency Policy (CP) ~\cite{prasad2024consistency}, AdaFlow~\cite{hu2024adaflow}, and the recently proposed Vision-to-Action flow matching policy (VITA)~\cite{gao2025vita}; 
(iii) 3D Point Cloud-based Policies: including 3D Diffusion Policy (DP3)~\cite{ze20243d} and its streamlined variant (Simple DP3), as well as contemporary SOTA spatial-aware methods such as $\mathrm{H}^3\mathrm{DP}$~\cite{lu2025h} and FreqPolicy~\cite{su2025freqpolicy}. 

\subsubsection{Observations and text queries.}
Each policy receives synchronized RGB images, depth-reconstructed point clouds (with calibrated camera intrinsics/extrinsics), and proprioception.
For Attention-DP3, the text query is a fixed task-level description of the manipulation target (e.g., the object name).
We use the same query for training and evaluation, avoiding test-time prompt tuning.

\subsubsection{Training and evaluation protocol.}
Following standard practice in sample-efficient imitation learning, we train all methods with 10 expert demonstrations per task, except on DexArt, where we use 100 demonstrations because of the complexity of articulated-object manipulation. We optimize all models with AdamW~\cite{loshchilov2017decoupled} using a learning rate of $10^{-4}$ ($\beta=(0.95,0.999)$, $\epsilon=10^{-8}$, weight decay $10^{-6}$), a batch size of 128, and no gradient accumulation. We train for 1000 epochs on MetaWorld and for 3000 epochs on the other benchmarks. We assess 20 episodes every 200 epochs and calculate the mean of
the top five success rates. To evaluate clutter generalization, we introduce distractor objects with progressively increasing severity to create cluttered conditions unseen during training, while keeping the task goal unchanged (Sec.~\ref{subsec:clutter}). All implementations are built in PyTorch and evaluated on a single NVIDIA RTX 3090 GPU.

\subsection{Results on Simulation Benchmarks}
\label{subsec:main_results}

Table~\ref{tab:adroit_dexart_results} and Table~\ref{tab:metaworld_main_with_tasks} present the quantitative comparison of Attention-DP3 against strong baselines. 
In the highly challenging MetaWorld suite, our Attention-DP3 achieves an average success rate of 0.73, which is higher than the score of DP3 and VITA. On spatial-reasoning-intensive tasks such as \textit{Push-Wall} and \textit{Pick-Place}, Attention-DP3 provides large gains of +0.43 and +0.42 over DP3, respectively, which indicates that the proposed semantic prior guides diffusion toward task-relevant geometric features. We observe similarly consistent improvements on Adroit (0.78 average versus 0.68 for DP3) and DexArt (0.56 average versus 0.52 for DP3), confirming that the benefit of semantic attention extends to both dexterous manipulation and articulated-object manipulation.

\begin{table}[t]
\centering
\small
\setlength{\tabcolsep}{5pt}
\renewcommand{\arraystretch}{1.15}
\caption{\textbf{Success Rates on Adroit and DexArt Benchmarks.} Missing entries (--) indicate methods not evaluated on that benchmark.}
\label{tab:adroit_dexart_results}
\resizebox{\columnwidth}{!}{%
\begin{tabular}{lccc|c|cccc|c}
\toprule
\multirow{2}{*}{\textbf{Method}} &
\multicolumn{3}{c|}{\textbf{Adroit}} & \multirow{2}{*}{\textbf{Avg}} &
\multicolumn{4}{c|}{\textbf{DexArt}} & \multirow{2}{*}{\textbf{Avg}} \\
\cmidrule(lr){2-4} \cmidrule(lr){6-9}
& Door & Pen & Hammer & & Laptop & Faucet & Toilet & Bucket & \\
\midrule
IBC~\cite{florence2022implicit}         & 0.00 & 0.00 & 0.09 & 0.03 & 0.03 & 0.07 & 0.00 & 0.14 & 0.06 \\
BCRNN~\cite{mandlekar2021matters}       & 0.00 & 0.00 & 0.09 & 0.03 & 0.03 & 0.01 & 0.00 & 0.05 & 0.02 \\
AdaFlow~\cite{hu2024adaflow}    & 0.27 & 0.18 & 0.45 & 0.30 & --   & --   & --   & --   & --   \\
CP~\cite{prasad2024consistency}          & 0.31 & 0.13 & 0.45 & 0.30 & --   & --   & --   & --   & --   \\
$\mathrm{H}^3\mathrm{DP}$~\cite{lu2025h} & --   & --   & --   & --   & 0.81 & 0.34 & 0.70 & \cellcolor{blue!15}\textbf{0.28} & 0.53 \\
FreqPolicy~\cite{su2025freqpolicy}  & --   & --   & --   & --   & \cellcolor{blue!15}\textbf{0.85} & 0.30 & \cellcolor{blue!15}\textbf{0.77} & 0.25 & 0.54 \\
DP~\cite{chi2025diffusion}          & 0.37 & 0.13 & 0.45 & 0.32 & 0.69 & 0.23 & 0.58 & 0.20 & 0.43 \\
DP3~\cite{ze20243d}         & 0.62 & 0.43 & \cellcolor{blue!15}\textbf{1.00} & 0.68 & 0.77 & 0.36 & 0.70 & 0.25 & 0.52 \\
Simple DP3~\cite{ze20243d}  & 0.58 & 0.46 & \cellcolor{blue!15}\textbf{1.00} & 0.68 & 0.79 & 0.26 & 0.63 & 0.22 & 0.48 \\
VITA~\cite{gao2025vita}        & 0.81 & \cellcolor{blue!15}\textbf{0.55} & 0.96 & 0.77 & 0.82 & \cellcolor{blue!15}\textbf{0.39} & 0.72 & \cellcolor{blue!15}\textbf{0.28} & 0.55 \\
\textbf{Ours} & \cellcolor{blue!15}\textbf{0.83} & 0.50 & \cellcolor{blue!15}\textbf{1.00} & \cellcolor{blue!15}\textbf{0.78} & \cellcolor{blue!15}\textbf{0.85} & 0.37 & 0.72 & \cellcolor{blue!15}\textbf{0.28} & \cellcolor{blue!15}\textbf{0.56} \\
\bottomrule
\end{tabular}%
}
\end{table}

\begin{table}[t]
\centering
\caption{\textbf{MetaWorld results by difficulty with representative tasks.}
For each difficulty, we show two representative tasks plus an ellipsis indicating additional tasks in the same split, and report the difficulty-average (Avg) over all tasks in that split; Overall is the mean over all MetaWorld tasks.
Full per-task results are provided in the supplementary material.}
\label{tab:metaworld_main_with_tasks}
\small
\setlength{\tabcolsep}{3.0pt}
\renewcommand{\arraystretch}{1.08}
\resizebox{\columnwidth}{!}{%
\begin{tabular}{lcccc|cccc|cccc|cccc|c}
\toprule
\textbf{Method}
& \multicolumn{4}{c|}{\textbf{Easy}}
& \multicolumn{4}{c|}{\textbf{Medium}}
& \multicolumn{4}{c|}{\textbf{Hard}}
& \multicolumn{4}{c|}{\textbf{Very Hard}}
& \textbf{Overall} \\
\cmidrule(lr){2-5}\cmidrule(lr){6-9}\cmidrule(lr){10-13}\cmidrule(lr){14-17}
& handle-pull & handle-pull-side & $\cdots$ & \textbf{Avg}
& push-wall & hammer & $\cdots$ & \textbf{Avg}
& pick-place & push & $\cdots$ & \textbf{Avg}
& pick-place-wall & disassemble & $\cdots$ & \textbf{Avg}
& \\
\midrule
DP
& 0.27 & 0.23 & -- & 0.836
& 0.20 & 0.15 & -- & 0.311
& 0.00 & 0.30 & -- & 0.090
& 0.05 & 0.43 & -- & 0.266
& 0.574 \\
DP3
& 0.34 & 0.53 & -- & 0.857
& 0.49 & 0.76 & -- & 0.525
& 0.12 & 0.51 & -- & 0.248
& 0.35 & 0.69 & -- & 0.442
& 0.669 \\
VITA
& 0.20 & 0.52 & -- & 0.839
& 0.49 & 0.90 & -- & 0.546
& 0.15 & 0.70 & -- & 0.315
& 0.48 & 0.61 & -- & 0.548
& 0.683 \\
\midrule
Ours
& \cellcolor{blue!15}\textbf{0.46} & \cellcolor{blue!15}\textbf{0.62} & -- & \cellcolor{blue!15}\textbf{0.870}
& \cellcolor{blue!15}\textbf{0.92} & \cellcolor{blue!15}\textbf{0.95} & -- & \cellcolor{blue!15}\textbf{0.599}
& \cellcolor{blue!15}\textbf{0.54} & \cellcolor{blue!15}\textbf{0.82} & -- & \cellcolor{blue!15}\textbf{0.378}
& \cellcolor{blue!15}\textbf{0.60} & \cellcolor{blue!15}\textbf{0.86} & -- & \cellcolor{blue!15}\textbf{0.620}
& \cellcolor{blue!15}\textbf{0.726} \\
\bottomrule
\end{tabular}%
}
\end{table}

\subsection{Real-world Experiment Results}
\begin{figure}[h!]
    \centering
    \includegraphics[width=0.9\linewidth]{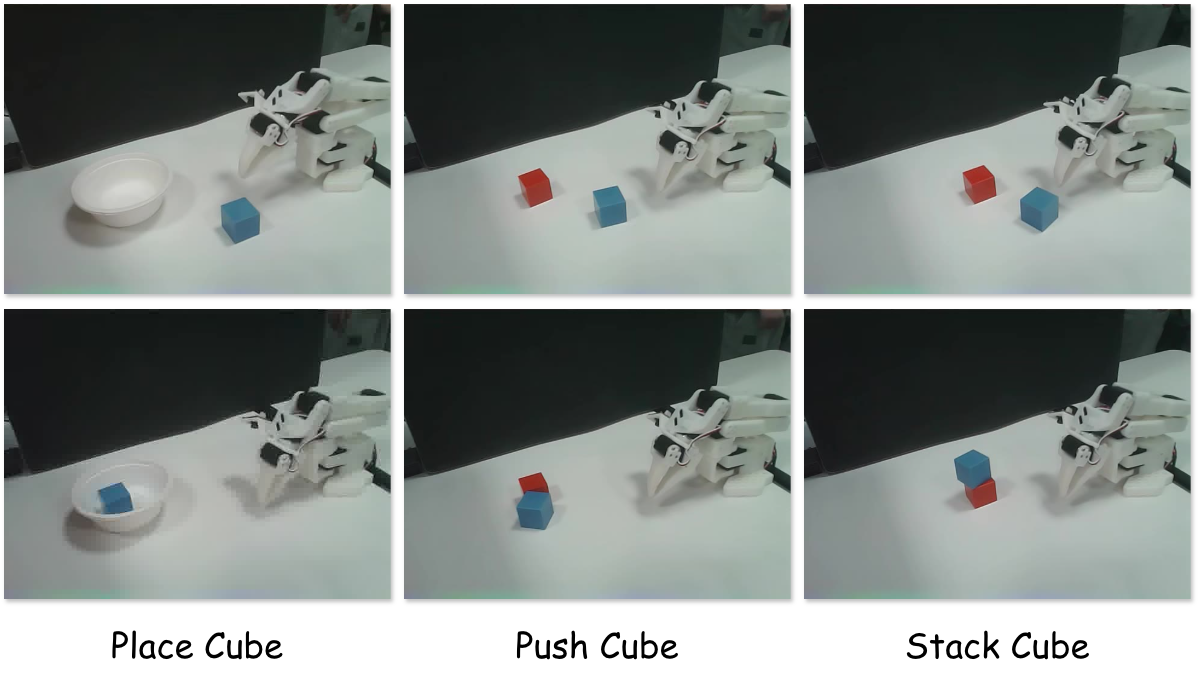}
    \caption{\textbf{Real-world experimental setting on SO101.} We evaluate three tasks, including Place Cube, Push Cube, and Stack Cube.}
    \label{fig:real_world_setting}
\end{figure}

We further validate Attention-DP3 on real-robot SO101 platform. Figure~\ref{fig:real_world_setting} shows the real-world experiment setup. We evaluate three representative tabletop tasks. In \textit{Place Cube}, the robot places the cube into the bowl. In \textit{Push Cube}, success is achieved when the blue cube is pushed to within 2 cm of the red cube. In \textit{Stack Cube}, success is achieved when the blue cube is placed stably on top of the red cube. We report success rates over repeated 20 rollouts under the same evaluation protocol for both methods.

As shown in Table~\ref{tab:real_world_results}, Attention-DP3 improves all three tasks with the largest gain on \textit{Push Cube}, which requires precise spatial reasoning under local contact changes. The average success rate improves from 0.52 to 0.73, which supports that geometry-aligned attentional conditioning transfers effectively from simulation to real-world manipulation.

% \begin{table}[t]
% \centering
% \caption{\textbf{Real-world manipulation results on SO101.} Attention-DP3 consistently outperforms DP3 across all tasks.}
% \label{tab:real_world_results}
% \resizebox{\columnwidth}{!}{%
% \begin{tabular}{lcc}
% \toprule
% \textbf{Task} & \textbf{Attention-DP3} & \textbf{DP3} \\
% \midrule
% Stack Cube & 0.70 & 0.50 \\
% Push Cube & 0.65 & 0.30 \\
% Place Cube & 0.85 & 0.75 \\
% \midrule
% \textbf{Average} & \textbf{0.73} & \textbf{0.52} \\
% \bottomrule
% \end{tabular}%
% }
% \end{table}

\begin{table}[t]
\centering
\caption{\textbf{Real-world results on SO101.} Attention-DP3 consistently outperforms DP3 across all tasks.}
\label{tab:real_world_results}
\small
\begin{tabular}{lcccc}
\toprule
\textbf{Task} & Stack Cube & Push Cube & Place Cube & \textbf{Average} \\
\midrule

\textbf{DP3} & 0.50 & 0.30 & 0.75 & 0.52 \\
\textbf{Attention-DP3} & \textbf{0.70} & \textbf{0.65} & \textbf{0.85} & \textbf{0.73} \\
\bottomrule
\end{tabular}
\end{table}

\subsection{Robustness to Visual Clutter}
\label{subsec:clutter}

We evaluate visual-clutter robustness in both simulation and real-world settings.

% Preamble

% In body
\begin{figure}[t]
    \centering
    \begin{subfigure}[t]{0.45\linewidth}
        \centering
        \includegraphics[width=\linewidth]{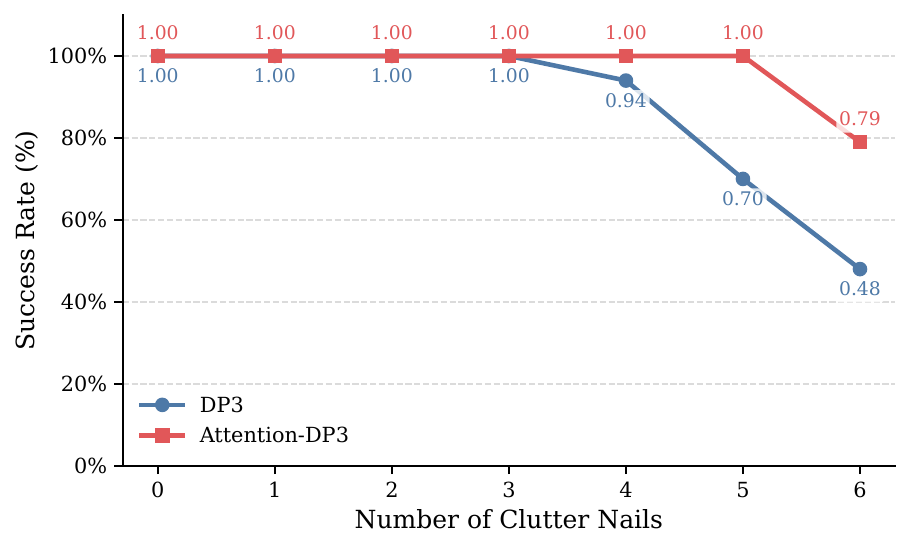}
        \caption{\textbf{Adroit Hammer (Clutter: Nails).} Success rate under increasing distractor nails.}
        \label{fig:adroit_clutter_pdf}
    \end{subfigure}\hfill
    \begin{subfigure}[t]{0.45\linewidth}
        \centering
        \includegraphics[width=\linewidth]{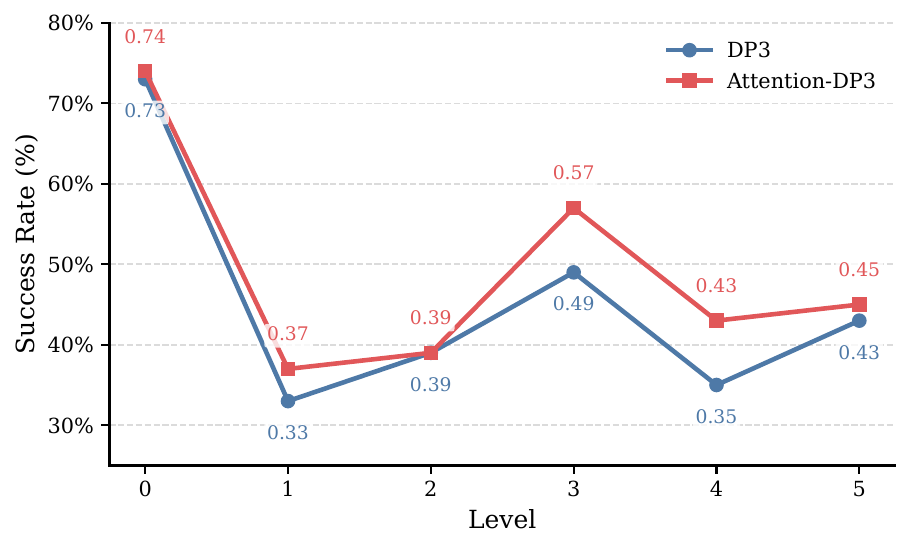}
        \caption{\textbf{MetaWorld Stick-Push (Clutter: Blocks).} Success rate under increasing distractor blocks.}
        \label{fig:metaworld_clutter_pdf}
    \end{subfigure}
    \caption{\textbf{Robustness to visual clutter.} Our attention mechanism improves robustness under unseen distractors across both tasks.}
    \label{fig:clutter_robustness}
\end{figure}

\begin{figure}[t]
    \centering
    \includegraphics[width=0.85\linewidth]{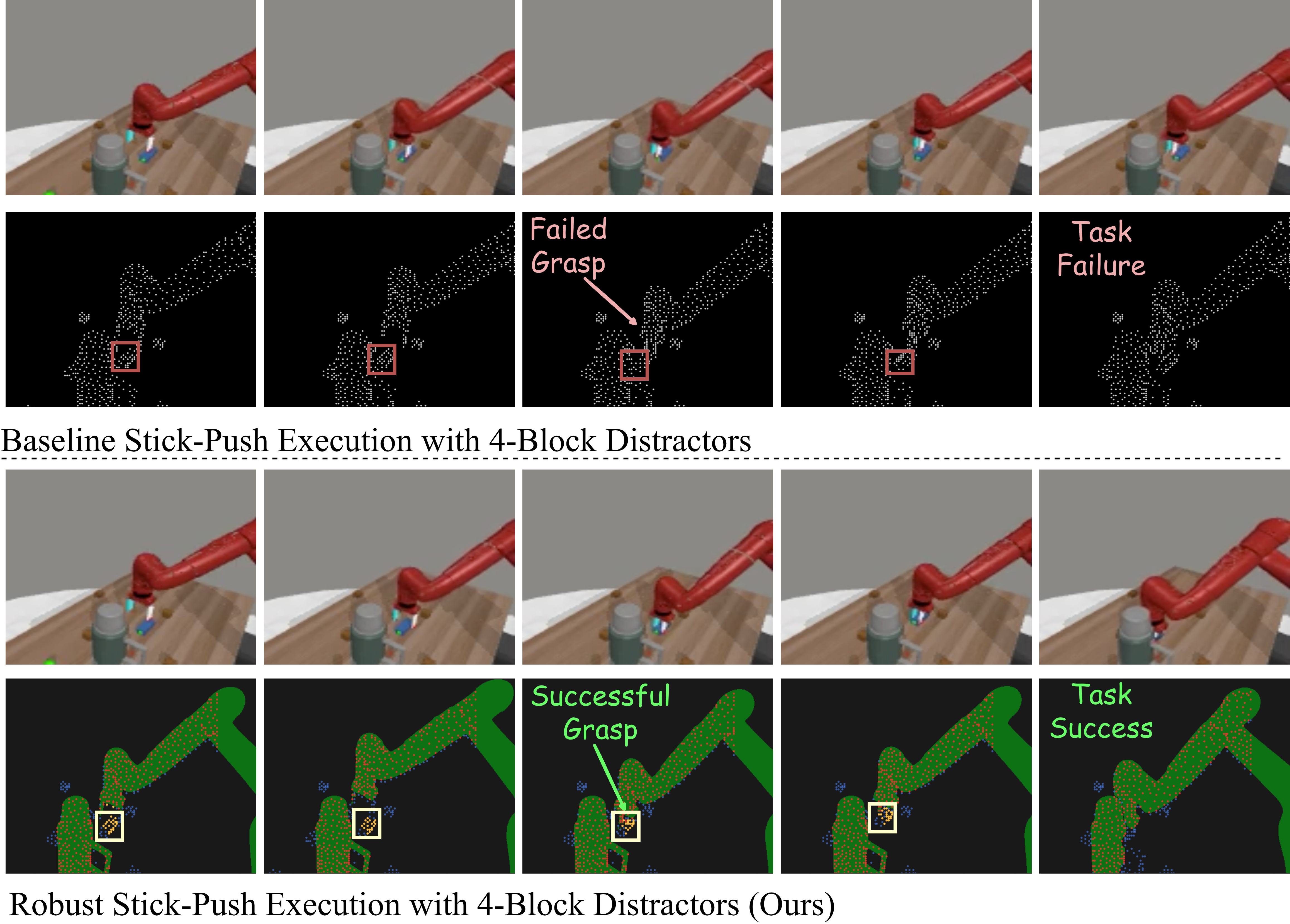}
    \caption{\textbf{Qualitative comparison on MetaWorld Stick-Push with four distractor blocks.} Top: DP3; bottom: Attention-DP3.}
    \label{fig:clutter_outcome}
\end{figure}

To assess the robustness of the policies in cluttered environments, we design generalization scenarios characterized by progressively increasing levels of distraction. For the Adroit \textit{Hammer} task, we vary the number of distractor nails from 0 to 6. Similarly, for the MetaWorld \textit{Stick-Push} task, we define six distinct distraction levels based on the quantity of distractor blocks, ranging from 0 to 5. The 0-block setting represents a clean, distractor-free environment, while the subsequent levels correspond to the introduction of 1 to 5 unseen blocks randomly scattered across the functional workspace of the robot. Detailed configurations of these experimental setups are provided in the supplementary material.

As shown in Figure~\ref{fig:clutter_robustness}, on \textit{Adroit Hammer}, both methods are nearly unaffected by light clutter. However, their behaviors diverge significantly in the high-density regime (4--6 nails). DP3 exhibits a clear monotonic collapse, indicating its susceptibility to task-irrelevant geometry. In contrast, Attention-DP3 preserves a remarkably flat degradation curve, maintaining saturated success through moderate clutter and demonstrating substantially higher tolerance before failure.

A similar pattern is observed in the \textit{MetaWorld Stick-Push} task across varying density levels (0--5 blocks). As the number of blocks increases, the performance of DP3 exhibits fluctuations and a general downward trend, indicating unstable foreground extraction under distribution shifts. In contrast, Attention-DP3 consistently outperforms DP3 across almost all clutter levels. The most pronounced performance gains are observed at mid-to-high clutter levels (3--5 blocks), where distractors severely interfere with local geometric perception.

We further perform a real-world clutter stress test by progressively adding distractors to the SO101 scenes, including clips, sticks, and task-specific extra distractors. Table~\ref{tab:real_world_clutter_results} reports the average success rate across the three real-world tasks. While DP3 degrades substantially as clutter increases, Attention-DP3 maintains a higher success rate under all clutter settings, indicating that object-aware conditioning helps the policy preserve target focus in unstructured real-world scenes.

\begin{table}[t]
\centering
\caption{\textbf{Real-world clutter robustness on SO101.} Average success rate across Place Cube, Push Cube, and Stack Cube under progressively added distractors.}
\label{tab:real_world_clutter_results}
\small
\begin{tabular}{lcccc}
\toprule
\textbf{Method} & \textbf{Orig.} & \textbf{+Clip} & \textbf{+Stick} & \textbf{+Extra} \\
\midrule
DP3 & 0.52 & 0.43 & 0.38 & 0.13 \\
Attention-DP3 & \textbf{0.73} & \textbf{0.72} & \textbf{0.65} & \textbf{0.45} \\
\bottomrule
\end{tabular}
\end{table}

\section{Ablation Study}
\label{sec:ablation_and_analysis}

% We first analyze (i) the contribution of each attention field and (ii) the effect of the fusion strategy. We then evaluate system efficiency from the perspective of control bandwidth.

\subsection{Ablation on attention fields.}
Table~\ref{tab:ablation_fields} compares different field combinations: Targetness (T), Intra-target saliency (I), and Backgroundness (B). Single-field variants underperform, two-field variants show task-dependent trade-offs, and the complete three-field configuration yields the most consistent improvements and highest average, validating the full design.

\begin{table}[t]
\centering
\caption{\textbf{Ablation on semantic attention fields.}
We report success rates on Adroit \textit{Door} and MetaWorld tasks. \textbf{Avg.} is the mean over the four tasks.}
\label{tab:ablation_fields}
\scriptsize
\setlength{\tabcolsep}{2pt}
\renewcommand{\arraystretch}{0.88}
\begin{tabular}{l c c c c c c c c}
\toprule
\textbf{Variant} & \textbf{T} & \textbf{I} & \textbf{B} & \textbf{Door} & \textbf{Stick-Pull} & \textbf{Pick-Place} & \textbf{Push-Wall} & \textbf{Avg.} \\
\midrule
Full & $\checkmark$ & $\checkmark$ & $\checkmark$ & \cellcolor{blue!15}\textbf{0.83} & \cellcolor{blue!15}\textbf{0.59} & \cellcolor{blue!15}\textbf{0.54} & \cellcolor{blue!15}\textbf{0.92} & \cellcolor{blue!15}\textbf{0.72} \\
\midrule
T only & $\checkmark$ &  &  & 0.74 & 0.53 & 0.33 & 0.57 & 0.5425 \\
I only &  & $\checkmark$ &  & 0.73 & 0.51 & 0.32 & 0.64 & 0.55 \\
B only &  &  & $\checkmark$ & 0.68 & 0.50 & 0.33 & 0.63 & 0.535 \\
T+I & $\checkmark$ & $\checkmark$ &  & 0.80 & 0.30 & 0.39 & 0.58 & 0.5175 \\
T+B & $\checkmark$ &  & $\checkmark$ & 0.75 & 0.56 & 0.36 & 0.58 & 0.5625 \\
B+I &  & $\checkmark$ & $\checkmark$ & 0.82 & 0.54 & 0.38 & 0.64 & 0.595 \\
\bottomrule
\end{tabular}
\end{table}

\subsection{Ablation on fusion strategies.}
Table~\ref{tab:ablation_fusion} compares early fusion (concatenating attention and geometry at input) with late fusion. Late fusion performs best, especially when each field is encoded by an independent AttnEncoder before fusion with geometric features, indicating that field-wise encoding preserves complementary cues and reduces cross-field interference.
Since all fusion variants use the same lifted mask source, the gap is not explained by external semantics alone; separate geometry-aligned fields and late fusion are important for turning the semantic prior into effective 3D control.

\begin{table}[h!]
\centering
\caption{\textbf{Ablation on fusion strategy.}
We report success rates on Adroit \textit{Door} and MetaWorld tasks. \textbf{Avg.} is the mean over the four tasks.}
\label{tab:ablation_fusion}
\resizebox{\columnwidth}{!}{
\begin{tabular}{l l c c c c c}
\toprule
\textbf{Fusion Strategy} & \textbf{Encoder} & \textbf{Door} & \textbf{Stick-Pull} & \textbf{Pick-Place} & \textbf{Push-Wall} & \textbf{Avg.} \\
\midrule
Late fusion & 3 AttnEncoders + DP3Encoder & \cellcolor{blue!15}\textbf{0.83} & \cellcolor{blue!15}\textbf{0.59} & \cellcolor{blue!15}\textbf{0.54} & \cellcolor{blue!15}\textbf{0.92} & \cellcolor{blue!15}\textbf{0.72} \\
Late fusion & Single AttnEncoder + DP3Encoder & 0.69 & 0.42 & 0.36 & 0.62 & 0.5225 \\
Early fusion & Attn+PC concat $\rightarrow$ DP3 Encoder & 0.66 & 0.02 & 0.07 & 0.27 & 0.255 \\
\bottomrule
\end{tabular}}
\end{table}

\subsection{Inference and Operational Latency}
We analyze the operational bandwidth of Attention-DP3 via control latency.
To isolate the overhead from semantic attention branches under asynchronous execution, we define the \textit{effective control step latency} $L_{step}$ as the average time per executed action step.
With prediction horizon $H=16$ and execution window $k=8$, each inference cycle spans $T_{phys}=0.8$\,s of physical execution at 10\,Hz.
We compute
\begin{equation}
L_{step}=\frac{T_{phys}+T_{inf}}{k},
\end{equation}
where $T_{inf}$ is the end-to-end inference time, including the Grounded-SAM-2 (GS2) perception pipeline.

Table~\ref{tab:system_bandwidth} shows that Attention-DP3 preserves smooth control bandwidth. Its mean $L_{step}$ is 0.198\,s, versus 0.122\,s for DP3, a modest increase without visible stuttering in asynchronous deployment. The policy-side overhead is minimal: compared with DP3, Attention-DP3 adds only 0.001\,s per step, 0.01\,GB peak memory, and 3.8M parameters. The remaining end-to-end cost comes from the frozen GS2 perception branch, whose lighter Swin-T variant reduces perception latency from 0.100\,s to 0.044\,s.

\begin{table}[t]
\centering
\caption{\textbf{Effective Control Step Latency.} Per-step latency $L_{step}$ for execution window $k=8$ ($T_{phys}=0.8$\,s), averaging physical execution and inference latency $T_{inf}$. Measurements use one NVIDIA RTX 3090 GPU.}
\label{tab:system_bandwidth}
\small
\begin{tabular}{l cc | c}
\toprule
\textbf{Task} & \textbf{DP3 (Baseline)} & \textbf{Attn-DP3 (Ours)} & \textbf{Latency Gap} \\
\midrule
Adroit Pen    & 0.120 s & 0.213 s & \textcolor{teal}{+0.093 s} \\
Adroit Door   & 0.124 s & 0.174 s & \textcolor{teal}{+0.050 s} \\
Adroit Hammer & 0.123 s & 0.209 s & \textcolor{teal}{+0.086 s} \\
\midrule
\textbf{Mean} & \textbf{0.122 s} & \textbf{0.198 s} & \textbf{+0.076 s} \\
\bottomrule
\end{tabular}
\end{table}

\subsection{Diagnostic Analysis of Failure Modes}
We evaluate whether failures of Attention-DP3 are primarily caused by 2D grounding errors. We consider three tasks: Coffee-Push, Peg-Unplug-Side, and Reach-Wall. For each task, we select 10 failure episodes of Attention-DP3 and rerun DP3 using identical random seeds.
Table~\ref{tab:failure_deep_diagnosis} shows that most Attention-DP3 failures in Coffee-Push and Peg-Unplug-Side come from missed detections by Grounding DINO (8/10 and 7/10), while DP3 also fails frequently on the same episodes (7/8 and 5/7), indicating inherently difficult cases rather than Attention-DP3-specific failures. In Reach-Wall, grounding is always correct (0/10), yet DP3 fails in all episodes (10/10), suggesting physical feasibility as the bottleneck. Thus, grounding errors explain many cluttered-task failures, but correct grounding does not eliminate physical constraints.

\begin{table}[ht]
\centering
\caption{\textbf{Diagnostic Analysis of Failure Mode Overlap.} We isolate 10 Attention-DP3 failure episodes and rerun DP3 with identical seeds. "Overlap Fail" reports DP3 failures when Attention-DP3 suffers Grounding DINO errors.}
\label{tab:failure_deep_diagnosis}
\resizebox{\columnwidth}{!}{
\begin{tabular}{l c c c c}
\toprule
\textbf{Task} & \textbf{Total Failures} & \textbf{Attn-DP3 (Fail via DINO)} & \textbf{DP3 (Overlap Fail)} & \textbf{DP3 (Total Fail)} \\
\midrule
\textit{Coffee-Push}     & 10 & 8 & 7 / 8 & 9 / 10 \\
\textit{Peg-Unplug-Side} & 10 & 7 & 5 / 7 & 8 / 10 \\
\textit{Reach-Wall}      & 10 & 0 & N/A   & 10 / 10 \\
\bottomrule
\end{tabular}}
\end{table}

\subsection{Robustness to Imperfect 2D Perception}
We further examine whether Attention-DP3 depends critically on accurate 2D segmentation. For each of five tasks, we evaluate 100 rollouts with matched initial seeds for Attention-DP3 and DP3, and categorize each episode by Attention-DP3 key-step segmentation quality. With incorrect segmentation, Attention-DP3 remains comparable to DP3 (29/134 vs. 30/134 successful rollouts), suggesting that erroneous masks do not introduce disproportionate failures. With correct segmentation, Attention-DP3 substantially outperforms DP3 (233/366 vs. 155/366), indicating that reliable object cues are effectively exploited. Direct mask perturbation shows the same trend: replacing predicted masks with ground-truth masks only improves average success from 0.524 to 0.542, while random mask dropout degrades smoothly and remains above DP3 even at $p=0.9$ (0.456 vs. 0.370). We also test prompt sensitivity with five generated prompts per task; generated and manual prompts perform comparably (0.73 vs. 0.75 mIoU, 0.55 vs. 0.56 success rate), suggesting that prompts mainly specify the target object rather than drive policy gains.

\subsection{Robustness to Perceptual Foundation Models}
To test robustness to the perception backbone, we vary the grounding model used by Attention-DP3. Concretely, we replace the default Grounding DINO Swin-B with a lightweight Swin-T variant, denoted Attention-DP3 (Swin-T), and compare it with the standard Attention-DP3 and the DP3 baseline. As shown in Table~\ref{tab:robustness_perceptual}, downsizing the grounding model yields comparable success rates despite reduced detection precision. Notably, Attention-DP3 (Swin-T) reaches 100\% success on Hammer, outperforming both the standard model and DP3. On spatial reasoning tasks such as Push-Wall and Pick-Place, it remains markedly stronger than the geometry-only baseline. Overall, these results suggest that our semantic-aware design is resilient to perceptual noise and does not rely on large or expensive vision backbones to achieve strong manipulation performance.

\begin{table}[ht]
\centering
\caption{\textbf{Robustness to Perceptual Foundation Models.} Success rates of DP3, Attention-DP3, and Attention-DP3 (Swin-T) using a lightweight Grounding DINO Swin-T backbone.}
\label{tab:robustness_perceptual}
\scriptsize
\setlength{\tabcolsep}{2pt}
\renewcommand{\arraystretch}{0.88}
\resizebox{\columnwidth}{!}{%
\begin{tabular}{lcccccc}
\toprule
\textbf{Method} & \textbf{Hammer} & \textbf{Box-Close} & \textbf{Push-Wall} & \textbf{Pick-Place} & \textbf{Shelf-Place} & \textbf{Stick-Pull} \\
\midrule
DP3 & 0.76 & 0.42 & 0.49 & 0.12 & 0.17 & 0.27 \\
Attn-DP3 (Swin-B) & 0.95 & 0.52 & 0.92 & 0.54 & 0.22 & 0.59 \\
Attn-DP3 (Swin-T) & \textbf{1.00} & 0.43 & 0.84 & 0.47 & 0.17 & 0.53 \\
\bottomrule
\end{tabular}
}
\end{table}

\section{Conclusion}
We presented Attention-DP3, a spatially object-aware 3D diffusion policy that mitigates clutter-induced perceptual ambiguity without altering the DP3 diffusion formulation. By lifting open-vocabulary 2D masks into geometry-aligned 3D object priors and applying Tri-Field Attentional Conditioning, it reliably binds language-specified targets to their 3D extents and remains robust under occlusion and heavy distraction. Experiments on Adroit, DexArt, MetaWorld, and the real-world SO101 platform show consistent gains over DP3, and clutter stress tests confirm strong zero-shot stability as distractors scale. We hope this work highlights lightweight, geometry-preserving object-level prompting as a practical route to robust manipulation in unstructured environments.

\section{Acknowledgment}
This paper is supported by the National Natural Science Foundation of China (62422610, U23A20386, 62441231, 62276045, 62576072), Liao Ning Science and Technology Plan (2025JH2/101330121, 2025JH2/101330124, 2023JH26/10200016), and Dalian City Science and Technology Innovation Fund (2023JJ11CG001).

\clearpage

\bibliographystyle{splncs04}
\bibliography{main}
\end{document}